\documentclass[a4paper, 10pt, conference]{ieeeconf}   
\usepackage{FG2025}
\usepackage{graphicx}
\usepackage{booktabs}
\usepackage{caption}
\usepackage{siunitx}
\usepackage{microtype}

\FGfinalcopy 

\IEEEoverridecommandlockouts                              
\def\FGPaperID{****} 

\title{\LARGE \bf
EdgeEar: Efficient and Accurate Ear Recognition for Edge Devices
}

\author{\parbox{16cm}{\centering
    {\large Camile Lendering$^{1,2*}$, Bernardo Perrone Ribeiro$^{1,2*}$, \v{Z}iga Emer\v{s}i\v{c}$^{1}$, and Peter Peer$^{1}$}\\
    {\normalsize
    $^1$ Faculty of Computer and Information Science, University of Ljubljana, Ve\v{c}na pot 113, 1000 Ljubljana\\
    $^2$ Universitat Pompeu Fabra, Tànger, 122-140 08018 Barcelona}}
    \thanks{$^*$ Equal contribution.\
    This research was supported by the ARIS Research Programme P2-0214 ``Computer Vision''.}
}
\begin{document}

\ifFGfinal
\thispagestyle{empty}
\pagestyle{empty}
\else
\author{Anonymous FG2025 submission\\ Paper ID \FGPaperID \\}
\pagestyle{plain}
\fi
\maketitle

\begin{abstract}

Ear recognition is a contactless and unobtrusive biometric technique with applications across various domains. However, deploying high-performing ear recognition models on resource-constrained devices is challenging, limiting their applicability and widespread adoption. This paper introduces \textit{EdgeEar}, a lightweight model based on a proposed hybrid CNN-transformer architecture to solve this problem. 
By incorporating low-rank approximations into specific linear layers, EdgeEar reduces its parameter count by a factor of $50$ compared to the current state-of-the-art, bringing it below two million while maintaining competitive accuracy. 
Evaluation on the Unconstrained Ear Recognition Challenge (UERC2023) benchmark shows that EdgeEar achieves the lowest EER while significantly reducing computational costs.
These findings demonstrate the feasibility of efficient and accurate ear recognition, which we believe will contribute to the wider adoption of ear biometrics.

\end{abstract}

\section{INTRODUCTION}

Ear recognition has gained attention as a reliable biometric modality for unobtrusive and contactless authentication~\cite{emersic2017}. Despite advances in deep learning yielding significant performance improvements, state-of-the-art models often rely on computationally intensive architectures, making them unsuitable for resource-constrained edge devices and limiting its wider use on general hardware. Addressing this gap is crucial for secure and efficient authentication across mobile, IoT, and various real-world applications.

Building on the success of lightweight face recognition models such as EdgeFace~\cite{George_IEEETBIOM_2024}, we introduce EdgeEar, a model designed for ear recognition on edge devices. With fewer than $2$ million parameters, EdgeEar balances accuracy and efficiency by adapting EdgeFace’s hybrid CNN-transformer architecture to open-set ear recognition.

\begin{figure}[ht!]
    \centering
    \includegraphics[width=1\columnwidth]{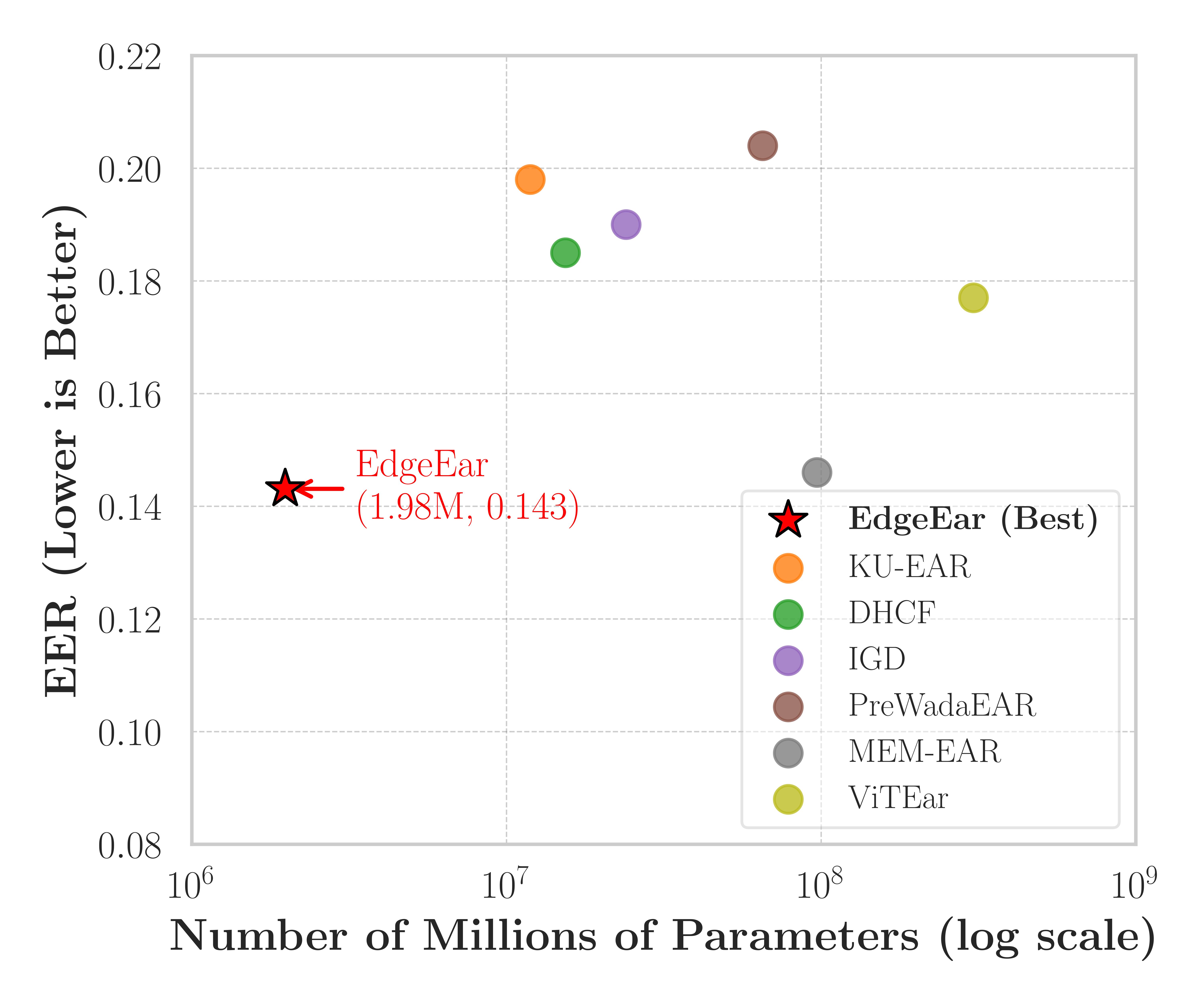}
    \caption{
    Comparison plot of Equal Error Rate (EER) versus the number of parameters (in M) for each UERC2023 approach. EdgeEar achieves a new state-of-the-art EER of $0.143$, surpassing the previous best model's EER of $0.146$, while maintaining a significantly smaller parameter count of $1.98M$ compared to $97.0M$.
    }
    \label{fig:eer_v_modelsize}
\end{figure}

Extensive evaluation on the Unconstrained Ear Recognition Challenge 2023 (UERC2023) benchmark~\cite{uerc2023} demonstrates that EdgeEar achieves performance comparable to state-of-the-art deep learning architectures across key metrics, such as Equal Error Rate, as shown in Fig. \ref{fig:eer_v_modelsize}, while maintaining a significantly smaller parameter count. These findings highlight the potential of adapting edge-optimized face recognition architectures for ear recognition in resource-constrained scenarios. Key contributions of this work include:
\begin{itemize} 
    \item We introduce \textit{EdgeEar}, the first ear recognition model specifically designed for edge devices. Combining a hybrid CNN-Transformer architecture with fewer than two million parameters, \textit{EdgeEar} achieves state-of-the-art performance on the UERC2023 benchmark, attaining an Equal Error Rate (EER) of $0.143$, Area Under Curve (AUC) of $0.904$, and Rank-1 (R1) accuracy of $0.929$, while maintaining high computational efficiency.
    
    \item We apply Low Rank Linear (LoRaLin) layers \textit{selectively} within the attention modules of the architecture. Ablation study demonstrates that this approach reduces model complexity and computational cost while maintaining recognition accuracy.
\end{itemize}

\section{Related work}

\subsection{Ear Biometrics and Core Challenges} Biometric authentication utilizes unique physiological traits to verify identity, offering enhanced security and convenience over traditional key-, secret-, or token-based methods~\cite{Dargan_biometrics_survey}. The ear is particularly suitable for recognition due to its stable morphological features, which remain consistent from childhood to late adulthood~\cite{Ibrahim_effect_ears_age}. Unlike facial recognition, ear biometrics are less affected by expressions, aging, or accessories such as beards, mustaches, or glasses. Additionally, ear features are typically not obscured by hairstyles or cosmetic changes, ensuring reliable identification~\cite{Benzaoui_survey_earrec}.

However, ear recognition systems must function in unconstrained environments, where pose, ethnicity, gender, and occlusion can impair performance. This has pushed research toward stronger feature representations and larger datasets to handle real-world variability. The Unconstrained Ear Recognition Challenge (UERC)~\cite{uerc2023,uerc2019,uerc2017} revealed the strengths and limits of current methods, stressing the need for approaches that generalize to new identities and enable open-set recognition.

\subsection{Deep Learning Advances for Ear Recognition}
While early ear recognition systems relied on hand-crafted features, contemporary approaches primarily utilize deep learning architectures such as Convolutional Neural Networks (CNNs)~\cite{Emersic2019} and Vision Transformers (ViTs)~\cite{alexey_VIT} to learn data-driven embeddings~\cite{Ziga_ear_recog_review}. These models effectively extract discriminative features from large ear datasets, enhancing performance under various conditions.

For instance, Alshazly et al.~\cite{Alshazly_Ear_Resnet} improve ResNet through transfer learning to address data scarcity by employing feature extraction, fine-tuning, and SVM-based classification. Similarly, Korichi et al.~\cite{Korichi_TRICA} introduce TR-ICANet, which integrates CNN-based normalization, Independent Component Analysis (ICA), and tied-rank normalization to boost accuracy. However, both methods are evaluated only on closed-set recognition tasks, where all target identities are known during training, limiting their relevance for open-set scenarios.

\subsection{Lightweight CNN and Transformer Architectures}
As deep networks become increasingly complex, significant progress has been made in developing lightweight architectures tailored for memory- and power-constrained environments. MobileNet~\cite{Howard_mobilenet, Sandler_mobilenetv2} introduced depthwise and pointwise convolutions to reduce parameters and computational load. Other efficient CNNs—such as SqueezeNet~\cite{Iandola2_squeezenet}, ShuffleNet~\cite{Zhang_shufflenet}, ShiftNet~\cite{Wu_shift}, and GhostNet~\cite{Han_ghostnet}—further optimize efficiency through smaller kernels, channel splitting, and shifting techniques.

With the rise of Vision Transformers (ViTs), hybrid networks have been developed to integrate transformer capabilities with CNN efficiencies. However, the multi-head attention mechanism often remains a computational bottleneck. Models like Mobile-Former~\cite{Chen_mobileformer} and MobileViT~\cite{Mehta_mobilevit} partially address this issue, but still require higher Multiply-Adds (MAdds) and longer inference times.

EdgeNeXt~\cite{Maaz_edgenext} offers a solution by extending ConvNeXt~\cite{liu_convnext} with a Split Depth-wise Transpose Attention (SDTA) encoder, combining depth-wise convolutions, adaptive kernel sizes, and transpose attention across channels. This design significantly reduces MAdds compared to traditional self-attention, making EdgeNeXt suitable for mobile applications.

\subsection{Lightweight Ear Recognition}

While efficient biometric models have advanced in domains like face recognition, with solutions such as Idiap EdgeFace-XS~\cite{George_IEEETBIOM_2024} adapting EdgeNeXt to operate with fewer than two million parameters, lightweight ear recognition remains under-explored. Many existing ear recognition methods rely on closed-set protocols, assuming all target identities have been seen during training. For example, a recent model~\cite{mehta2024efficient} utilizes an ensemble of over $11$ million parameters, achieving $98.74\%$ accuracy on the IITD-II dataset but only within a closed-set framework. Similarly, MobileNet-based approaches~\cite{xu2022efficient} employ $3.5$ to $5.4$ million parameters, which may be too large for memory-constrained devices and are limited by closed-set evaluations.

To address these limitations, we introduce EdgeEar, a lightweight ear recognition model inspired by EdgeNeXt and EdgeFace, with fewer than two million parameters. This parameter efficiency aligns with the objectives of competitions like EFaR 2023~\cite{Kolf2023EFaR}, where compact architectures are evaluated for their performance-resource trade-off. Unlike prior methods, EdgeEar is evaluated in open-set scenarios, demonstrating robust performance and effective generalization to unseen identities. This development advances the feasibility of deploying ear recognition systems on resource-limited hardware.

\section{METHODS}

\subsection{EdgeEar Architecture}

\begin{figure*}[ht!]
    \centering
    \includegraphics[width=0.95\textwidth]{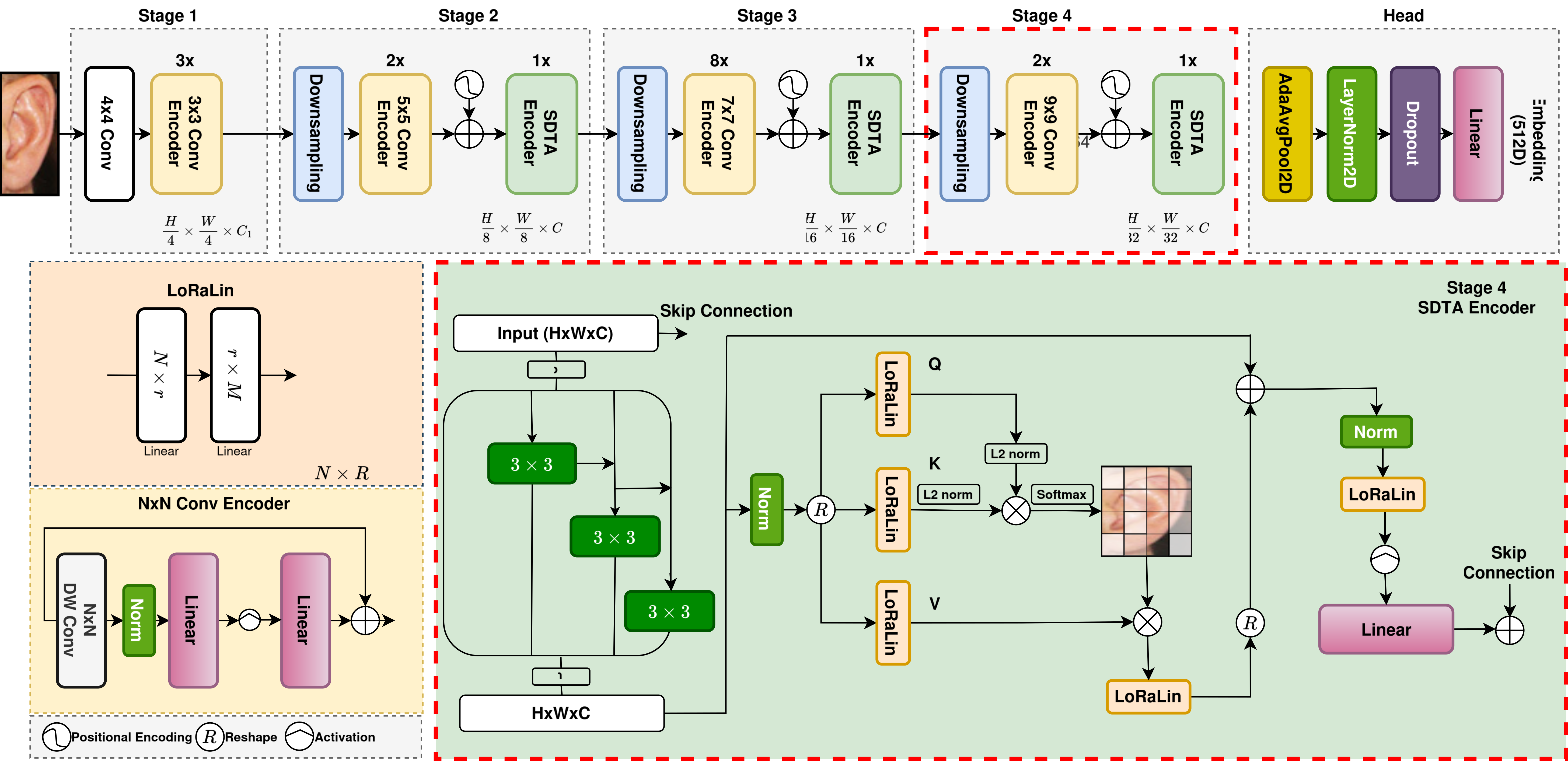}
    \caption{Schematic diagram of the proposed EdgeEar model, adapted from the EdgeFace~\cite{George_IEEETBIOM_2024} and EdgeNeXt~\cite{Maaz_edgenext} architectures. The modifications include \textit{selectively} applying \textit{LoRaLin} layers (highlighted in red) to three layers of the Stage~4 SDTA Encoder, while retaining full-rank linear layers for the remaining layers. }
    \label{fig:arch}
\end{figure*}

EdgeEar is a lightweight ear recognition model adapted from the \textit{EdgeFace} architecture~\cite{George_IEEETBIOM_2024} and built upon the \textit{EdgeNeXt} framework~\cite{Maaz_edgenext}. Designed for deployment on resource-constrained edge devices, EdgeEar maintains a hybrid CNN-Transformer structure, as illustrated in Fig.~\ref{fig:arch}.

EdgeEar modifies the original architecture by replacing specific linear layers in the Split Depth-wise Transpose Attention (SDTA) modules with Low Rank Linear (\textit{LoRaLin}) layers, as proposed in EdgeFace~\cite{George_IEEETBIOM_2024}. This reduces the total number of parameters while preserving the model's capacity to capture essential features for accurate ear recognition.

\subsection{Low Rank Linear Layers (\textit{LoRaLin})}

To improve parameter efficiency, EdgeEar integrates \textit{LoRaLin} layers into SDTA modules. Each linear transformation \( Y = W_{M \times N}X + b \) (input dimension \( N \), output dimension \( M \), bias \( b \)) is factorized as:

\begin{equation}
    Y = W_{M \times r}(W_{r \times N}X) + b,
\end{equation}

where \( W_{M \times r} \) and \( W_{r \times N} \) are low-rank matrices with rank \( r \), set by:

\begin{equation}
    r = \max\left(2, \gamma \cdot \min(M, N)\right).
\end{equation}

Here \( \gamma \) controls the rank ratio. This reduces parameters and computational complexity while enabling complexity-performance balancing through \( \gamma \).

\subsection{Selective Replacement Strategy}

In the final stage of the SDTA module (Stage~4 of the EdgeNeXt architecture), only the linear layers responsible for computing the query (Q), key (K), and value (V) projections are replaced with low-rank layers, following established low-rank approximation techniques~\cite{hu2022lora, wang2020linformer}. Specifically, the QKV linear layers are substituted with \textit{LoRaLin} layers using a rank ratio of $0.5$, while the remaining two linear layers adopt \textit{LoRaLin} layers with a rank ratio of $0.6$. This targeted modification balances parameter efficiency, maintaining full-rank convolutional encoders for feature extraction to capture global ear features. 

The resulting EdgeEar model produces a $512$-dimensional ear embedding and contains $1.98M$ parameters. Table~\ref{tab:edgeear_metrics} summarizes the computational metrics of the final model.

\begin{table}[htbp]
    \centering
    \caption{Computational Metrics of the EdgeEar Model.}
    \label{tab:edgeear_metrics}
    \begin{tabular}{@{}lc@{}}
        \toprule
        \textbf{Metric} & \textbf{Value} \\ 
        \midrule
        Number of Parameters                    & \num{1.98}~M \\
        Multiply-Adds Operations (MAdds) & \num{129.03}~M \\
        Floating Point Operations (FLOPs)       & \num{0.26}~GFLOPs \\
        \bottomrule
    \end{tabular}
\end{table}

\subsection{Implementation Details}

The model was trained in PyTorch using the AdamW optimizer (weight decay = $0.05$) with a cosine learning rate scheduler and a starting learning rate of \(3 \times 10^{-3}\). Cross Entropy with Label Smoothing of $0.1$ was employed as the loss function to enhance the model's generalization capabilities and mitigate overfitting~\cite{szegedy2016rethinking}. Xavier initialization was applied to the weights of all linear and convolutional layers to promote better convergence during training. Training was performed on a single NVIDIA RTX 4090 GPU with a batch size of $256$, over $1200$ epochs with early stopping.

Two datasets were used during training: 
\begin{enumerate}
    \item UERC2023~\cite{uerc2023}, containing $1310$ subjects with a total of $248.655$ samples.
    \item EARVN1.0~\cite{hoang2019earvn1}, containing $164$ subjects of Asian descent with a total of $28.412$ samples.
\end{enumerate}

All training images were augmented with Random Rotation, Color Jitter, Random Horizontal and Vertical Flips, Random Affine Transformation, Random Crop, and Gaussian Blur. The images were then resized to $(128 \times 128)$ pixels.

\subsection{Ablation Studies}
We conducted two categories of ablation studies:

\begin{enumerate}
    \item \textbf{Ablation With Non-Selective \textit{LoRaLin} with Varying \(\gamma\):} 
    To evaluate the effectiveness of the selective replacement strategy in the \textit{LoRaLin} layers, we conducted experiments where EdgeFace was instantiated with with varying \(\gamma\) while ensuring that the number of parameters remained below or near $2M$.
    \item \textbf{Ablation With Different Losses:} 
    We also performed experiments with different loss functions, such as regular cross entropy (CE) and ArcFace \cite{deng2019arcface}.
\end{enumerate}

\subsection{Evaluation}
\label{subsec:evaluation}

We evaluated our model on the sequestered dataset from the UERC 2023 competition for direct comparison with other architectures. Under an open-set protocol, we first generated $512$-dimensional embeddings for each pair of samples, computed cosine similarities between these embeddings, aggregated them per identity, and generated receiver operating characteristic (ROC) curves. By averaging these curves, we obtained a single representative ROC from which we derived key metrics: area under the curve (AUC), equal error rate (EER), False Non-Match Rate at 1\% False Match Rate – FNMR @ 1\% FMR (F1F), and Rank-1 accuracy. We also plotted the ROC curve by demographic subgroups (ethnicity and gender) to highlight bias considerations.

\section{RESULTS}

Table~\ref{tab:ablation-results} presents the results of our ablation study on EdgeNext variants with $\leq 2\,\mathrm{M}$ parameters. EdgeEar, trained with cross-entropy (CE) loss and label smoothing, achieves the lowest equal error rate (EER) of $0.143$, highest AUC of $0.904$, highest Rank-1 accuracy (R1) of $0.929$, and an F1F of $0.544$. 

Using ArcFace (margin: $0.2$, scale: $8$) instead of CE significantly worsens performance across all metrics, particularly increasing the EER to $0.229$ and lowering R1 to $0.571$, underscoring the suitability of CE loss for this task.

Removing label smoothing from CE leads to a trade-off: while EER increases to $0.198$, R1 and AUC slightly decrease to $0.927$ and $0.903$ respectively, with a decrease in F1F, suggesting that label smoothing improves overall metrics at the cost of higher F1F. Variants of EdgeFace, parameterized by $\gamma$, show strong AUC performance, with $\gamma = 0.6$ achieving the same AUC as EdgeEar ($0.904$), but none surpass EdgeEar in EER or overall balance across metrics. Notably, EdgeFace ($\gamma = 0.7$) achieves the lowest F1F ($0.305$), though at the expense of higher EER ($0.198$) and lower R1 ($0.879$).

\begin{table}[ht]
\centering
\caption{Ablation study results for model variants with $\leq\,2$M parameters. Metrics include Equal Error Rate (EER), Area Under Curve (AUC), Rank-1 accuracy (R1), and False Non-Match Rate at 1\% False Match Rate – FNMR @ 1\% FMR (F1F).}
\label{tab:ablation-results}
\begin{tabular}{@{}lcccc@{}}
\toprule
\textbf{Model Variant -- \# Params}                   & \textbf{EER} $\downarrow$ & \textbf{AUC} $\uparrow$ & \textbf{R1} $\uparrow$ & \textbf{F1F} $\downarrow$ \\ \midrule
EdgeEar (ours) -- $1.98M$                          & \textbf{0.143}             & \textbf{0.904}             & \textbf{0.929}            & $0.544$             \\
\quad - With ArcFace                     & $0.229$             & $0.850$             & $0.571$            & $0.707$             \\
\quad - CE loss w/o label smoothing      & $0.187$             & $0.903$             & $0.925$            & $0.359$             \\
EdgeFace ($\gamma = 0.5$) -- $1.51M$                & $0.202$             & $0.897$             & $0.884$            & $0.396$                 \\
EdgeFace ($\gamma = 0.6$) -- $1.77M$                & $0.191$             & \textbf{0.904}              & $0.907$            & $0.530$                 \\
EdgeFace ($\gamma = 0.7$) -- $2.03M$               & $0.198$              & $0.895$            & $0.879$         & \textbf{0.305}                 \\ 
\bottomrule
\end{tabular}
\end{table}

\begin{figure}[ht!]
    \centering
    \includegraphics[width=\columnwidth]{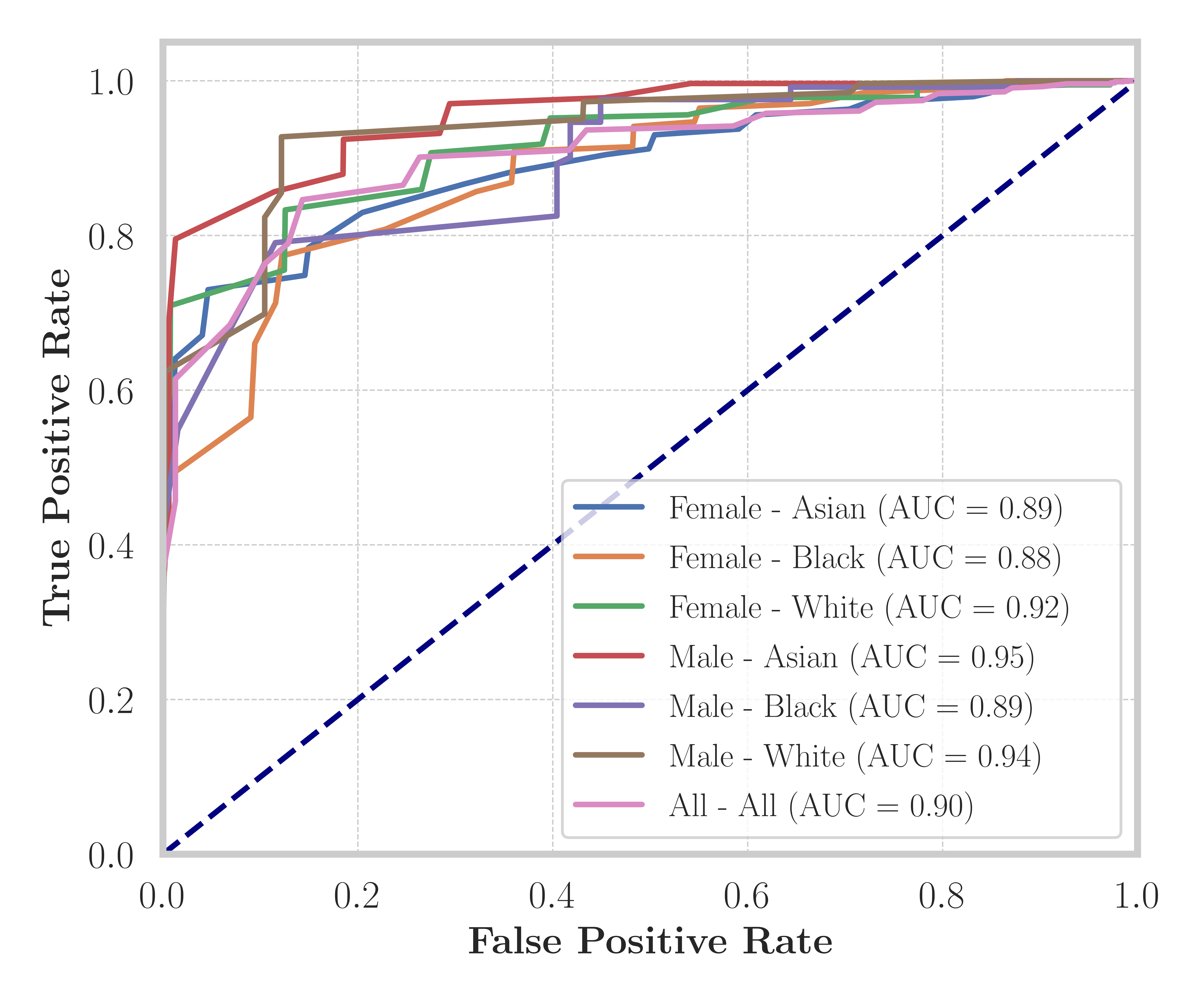}
    \caption{ROC curves for ethnicity and gender splits in the sequestered UERC23 dataset. Each curve represents the model's performance for a specific group, with the AUC annotated in the legend.}
    \label{fig:combined_roc}
\end{figure}

Figure~\ref{fig:combined_roc} shows the ROC curves for ethnicity and gender splits within the UERC23 dataset. EdgeEar achieves high AUC scores across all demographics, demonstrating robustness, though biases favoring Male-White and Male-Asian groups are evident. These biases stem from the overrepresentation of Male-White identities ($74.5\%$) in the training data and the inclusion of the EARVN1.0 dataset, which boosts Male-Asian representation.

\begin{table}[ht]
\centering
\caption{Comparison of EdgeEar with top-performing models from UERC2023. Metrics include Equal Error Rate (EER), Area Under Curve (AUC), False Non-Match Rate at 1\% False Match Rate – FNMR @ 1\% FMR (F1F), and the number of parameters.}
\label{tab:uerc_ranking}
\begin{tabular}{@{}lcccc@{}}
\toprule
\textbf{Model Variant}                   & \textbf{EER} $\downarrow$ & \textbf{AUC} $\uparrow$ & \textbf{F1F} $\downarrow$ & \textbf{\# Params} \\ \midrule
EdgeEar (ours)                           & \textbf{0.143}                 & $0.904$                 & $0.544$               & $1.98M$                  \\
MEM-EAR                & $0.146$                 & \textbf{0.915}                 & 0.313                & $97.0M$                 \\
ViTEar                 & $0.177$                 & $0.908$                 & \textbf{0.278}                & $304.9M$                 \\ 
\bottomrule
\end{tabular}
\end{table}

Table~\ref{tab:uerc_ranking} compares EdgeEar with top-performing models from UERC2023. EdgeEar achieves the lowest EER ($0.143$) while using $98.0\%$ fewer parameters than MEM-EAR and $99.3\%$ fewer parameters than ViTEar. Despite its compact size, EdgeEar maintains a competitive AUC ($0.904$) score.

These results demonstrate that EdgeEar delivers state-of-the-art performance while offering exceptional efficiency, making it a scalable solution for ear recognition tasks.

\addtolength{\textheight}{-0cm} 
\section{CONCLUSIONS}

In this paper, we present EdgeEar, a lightweight ear recognition model optimized for deployment on resource-constrained edge devices, with a parameter count of fewer than 2 million. While we did not formally participate in the UERC 2023 competition, we evaluated our model using the sequestered dataset and official evaluation protocol, enabling direct comparison with competition results. EdgeEar would rank first in EER all while operating at a significantly lower computational cost compared to other submissions.

Our findings suggest that model capacity is not the main limitation in achieving ear recognition performance comparable to other biometric modalities, such as face recognition. Instead, the size and diversity of datasets play a critical role. Future research should prioritize the development of larger and more diverse datasets to advance the field, enhance the applicability of ear-based biometrics, and mitigate demographic bias.

\section*{ETHICAL IMPACT STATEMENT}

\textit{EdgeEar} was developed using publicly available biometric datasets, with no additional data collection conducted. While the demographic diversity of these datasets is not guaranteed, we evaluate the model’s performance across known subsets to identify and address potential biases.
Safeguards should be implemented to prevent misuse such as in surveillance.

\newpage 
{\small
\bibliographystyle{ieee}
\bibliography{egbib}
}

\end{document}